\documentclass[3p,11pt,sort&compress]{elsarticle}

\usepackage{algorithm, algorithmic}
\usepackage{enumerate}
\usepackage{amsmath, amsthm, amssymb}
\usepackage{appendix}
\usepackage{graphicx}
\usepackage{wrapfig}
\usepackage{multirow}
\usepackage{hyperref}
\newcommand{\doiurl}[1]{DOI: \href{http://dx.doi.org/#1}{#1}}

\journal{International Journal of Applied Metaheuristic Computing, 3(4): 1-24, 2012 {\em [\doiurl{10.4018/jamc.2012100101}]}}

\begin{document}

\begin{frontmatter}

\title{Round-Table Group Optimization for Sequencing Problems}

\author{Xiao-Feng Xie}
\ead{xfxie@alumni.cmu.edu, xie@wiomax.com}

\address{The Robotics Institute, Carnegie Mellon University, Pittsburgh, PA 15213}

\begin{abstract}
\label{pap:abstract}

In this paper, a round-table group optimization (RTGO) algorithm is presented. RTGO is a simple metaheuristic framework using the insights of research on group creativity. In a cooperative group, the agents work in iterative sessions to search innovative ideas in a common problem landscape. Each agent has one base idea stored in its individual memory, and one social idea fed by a round-table group support mechanism in each session. The idea combination and improvement processes are respectively realized by using a recombination search (XS) strategy and a local search (LS) strategy, to build on the base and social ideas. RTGO is then implemented for solving two difficult sequencing problems, i.e., the flowshop scheduling problem and the quadratic assignment problem. The  domain-specific LS strategies are adopted from existing algorithms, whereas a general XS class, called socially biased combination (SBX), is realized in a modular form. The performance of RTGO is then evaluated on commonly-used benchmark datasets. Good performance on different problems can be achieved by RTGO using appropriate SBX operators. Furthermore, RTGO is able to outperform some existing methods, including methods using the same LS strategies. 

\end{abstract}


\end{frontmatter}

\section{Introduction}

Group creativity techniques, e.g., {\em brainstorming} \cite{Osborn1953}, have been widely studied in social science \cite{Nemeth:1986p980,Paulus:2000p1114}. A cooperative human group contains multiple individuals who have some interactions on ideas of each others \cite{Paulus:2000p1114}, and the group tries to find innovative solutions (or high-quality {\em ideas}) for a specific task by generating new (especially innovative) ideas spontaneously contributed by its members in iterative idea-generating sessions. 

Group creativity has been studied in both group and individual levels. At the group level, a {\em support mechanism} is used to assist individuals by providing useful external stimulus in their idea-generation process. The essential function of these mechanisms is to serve as {\em group memory} \cite{Betts2010,Satzinger1999} for the diffusion of innovative patterns \cite{Rogers2003} as well as providing diverse ideas \cite{Paulus:2000p1114}, based on a developing repository of nonprivate knowledge shared by members. For individuals, group memory might be heterogeneously shaped in {\em network structures} \cite{Lovejoy2010,Rulke2000}.

For each individual, the ideation process involves {\em idea selection} \cite{Rietzschel2010,Putman2009} and {\em idea generation} \cite{Nijstad2006,Paulus:2000p1114,Kohn2011}, based on its {\em individual memory} \cite{Glenberg:1997p1390,Newell:1972p1294}. A pre-selection mechanism \cite{Putman2009} might be used to pick out relevant ideas from both the individual and group memory. Then an idea-generation mechanism is used to build new idea(s) on selected ideas. Afterward, a post-selection mechanism \cite{Rietzschel2010} is applied to update the individual memory. Each individual also holds a {\em sharing mechanism} to contribute nonprivate knowledge \cite{Liu:2006p1045} for the group.

The idea-generation process is a critical part of the creative process. Since the research of Osborn \cite{Osborn1953}, {\em combine-and-improve ideas} 
has been a general brainstorming rule to form a single better idea by building on existing ideas. The total process of building on existing ideas might be divided into two parts, the idea combination process \cite{Kohn2011,Yu2011a} and the idea improvement process, roughly corresponding to the divergent and convergent processes, where the former is of paramount importance although the latter is also nontrivial \cite{Cropley2006}.

Human problem solving can be seen as searching in a structural space of states (ideas) \cite{Newell:1972p1294}. Thus, there is a natural similarity between creative idea-generating tasks and hard optimization problems \cite{Lovejoy2010}, both of them require efficiently achieving high-quality solutions.

Many optimization problems can be formulated as sequencing (or permutation) problems \cite{Koivisto2010,Starkweather:1991p2505}. All sequencing problems with $n$ nodes share the same problem space, in which each potential solution (or {\em state}) $\vec{x}$ is a permutation \{$x_{1}$, \dots, $x_{n}$\} of the integer values from 1 through $n$, only their objective functions $f(\vec{x})$ possess different structural properties. 

In this paper, we consider two typical sequencing problem examples, i.e., the quadratic assignment problem (QAP) \cite{Burkard:1997p2507,Taillard:1990p2487} and the flowshop scheduling problem (FSP) \cite{Beasley:1990p2481,Reeves:1995p2437,Taillard:1993p2472}, among some other examples \cite{Xie2012}. Both QAP and FSP are \textit{NP}-hard in the strong sense. FSP  is a well-known problem in intelligent manufacturing systems. QAP arises in many practical applications, e.g., design of grey patterns and website structure improvement \cite{Saremi:2008p2438}. QAP also serves as a generalization of some other important optimization problems \cite{Merz:2000p2516}. 

Exact algorithms, e.g., branch-and-bound \cite{Carlier:1996p2445}, can only be tractable for solving small-scale instances, whereas fast constructive heuristics \cite{Reeves:1985p2503,Nawaz:1983p2441} often obtain results that are far away from optimal. Thus, low-level search components, especially \textit{local search} (LS) and \textit{recombination search} (XS), as well as upper-level \textit{meta-heuristic frameworks}, have been integrated together for finding near optimal solutions within practical computational costs.

Each LS strategy improves an incumbent solution by intensively \textit{moves} based on some \textit{neighborhood search} operations \cite{Gambardella:1999p2521}, e.g., 2-opt. A LS strategy is defined as \textit{stable} if it only allows non-worse moves. Any stable LS, such as hill-climbing and the fast LS \cite{Gambardella:1999p2521}, cannot escape from the local minimum it first encounters. Some advanced LS strategies, such as simulated annealing \cite{Nearchou:2004p2443}, tabu search \cite{Nowicki:1996p2469,Taillard:1991p2451}, iterated local search \cite{Stutzle:1998p2467, Ruiz:2007p2496}, etc., incorporate some unstable moves so as to explore in a rugged problem landscape.

Each XS strategy generates a new solution by preserving positive clues in two parent solutions, which has an implicit advantage of adaptive leaping by utilizing the difference between two parents \cite{Merz:2000p2516}. Typical examples of XS operators for sequencing problems include order crossovers \cite{Murata:1996p2463,Starkweather:1991p2505}, LCS crossover \cite{Iyer:2004p2489}, similar block crossovers \cite{Ruiz:2006p2435}, partially mapped crossovers \cite{Starkweather:1991p2505}, distance preserving crossover \cite{Merz:2000p2516}, and path crossovers \cite{Ahuja:2000p2511}, etc.

Meta-heuristic frameworks, which use LS and XS strategies as their search components, have been applied for solving sequencing problems. Typical examples include ant colony optimization \cite{Gambardella:1999p2521,Maniezzo:1999p2461,Rajendran:2004p2504,Khalouli2011}, genetic algorithms and memetic algorithms \cite{Iyer:2004p2489, Merz:2000p2516, Murata:1996p2463, Reeves:1995p2437, Ruiz:2006p2435, Xu2011}, particle swarm optimization \cite{Deroussi:2005p2440,Tasgetiren:2007p2491}, and some other systems \cite{Xie:2009p1407}, etc.

In this paper, a round-table group optimizer (RTGO) is proposed for solving sequencing problems. We use the insights of previous research on group creativity to provide a theoretical context. RTGO is an extremely simple form of a cooperative group, in which each individual is an idea-generating agent. The agents work in iterative sessions to search innovative ideas in a common problem landscape. All agents are all intrinsically motivated to generate new ideas, and thus any negative group effects \cite{Paulus:2000p1114} are precluded. 

The group support mechanism is a simple round-table mechanism. In each session, the agents are randomly (re-)allocated around a round table, where each agent only accesses one {\em social idea} (in the group memory) from the adjacent neighbor. This simple mechanism captures a limited sense on communicative and cognitive interference \cite{Paulus:2000p1114}.

Each agent possesses only one {\em base idea} in its individual memory, and holds a XS strategy and a LS strategy respectively for the idea combination and improvement processes. The pre-selection mechanism only returns the base and social ideas for supporting socially biased learning \cite{Galef:1995p1128,MontesdeOca2011,Xie:2009p1407}. The XS strategy then generates an intermediate idea modified from the basic idea by combining stimulation cues in the social idea. Afterward, the LS strategy locally refines the intermediate one into a polished new idea. The post-selection mechanism then stores the new idea if it is not worse than the old base idea in its individual memory. For each agent, its individual memory always stores the best-so-far idea during its own search.

Our focus is then on examining the meta-heuristic framework, RTGO, with different idea combination processes. Two stable LS strategies respectively for FSP \cite{Ruiz:2006p2435,Ruiz:2007p2496} and QAP \cite{Gambardella:1999p2521} are chosen from prior work. In the ``big-valley'' structure \cite{Merz:2000p2516,Reeves:1995p2437} of sequencing problems, any stable LS strategies can only find near local optimal solutions, thus the capability of efficiently escaping from local minima can only be achieved by the idea combination process. In addition, it should be quite fair to compare the overall performance of RTGO to that of some existing optimization frameworks using commonly-used LS strategies. 

The remainder of this paper is organized as follows. In Section \ref{sec:RTGO}, RTGO is presented. In Section \ref{sec:Permutation}
, knowledge components for both sequencing problems are implemented into RTGO. In Section \ref{sec:Comparisons}, the characteristics of RTGO are investigated by performing computational experiments on some benchmark datasets. This paper is concluded in Section \ref{sec:Conclusion}.

\section{Round-Table Group Optimizer (RTGO)}
\label{sec:RTGO}

\begin{figure} [htbp]
\centering \includegraphics*[width=3.53in]{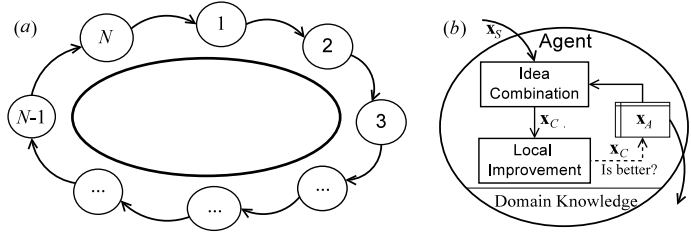}
\caption{(a) A round table seated with $N$ agents in a brainstorming group; and (b) the details of an agent.}
\label{Fig:RTGO}
\end{figure}

RTGO is a very simple meta-heuristic framework based on the insights of previous research on group creativity. As shown in Figure \ref{Fig:RTGO}, RTGO contains a small group of idea-generating agents that their interactions are supported by using a round-table mechanism. Each specific task is formulated into a problem landscape, where each state in the problem space is an {\em idea}. For the agents, the goal is to obtain a high-quality idea with a near optimal objective value in the problem space, through generating new ideas in iterative sessions. The actual algorithm performance will demonstrate the effectiveness of group creativity.
 
At the group level, a simple round-table mechanism is considered to provides background supports for the group memory among the agents, as shown in  Figure \ref{Fig:RTGO}(a). In each session, the agents are randomly assigned to the seats around a round table, and each agent shares a nonprivate idea in its individual memory as the only social idea for its clockwise neighbor. Afterward, agents can spontaneously generate new ideas using the base and social ideas. 

Each agent has a limited capability of generating innovative ideas based on available information, including available ideas and problem domain knowledge. Figure \ref{Fig:RTGO}(b) gives the details of an agent, which possesses basic characteristics of an individual in an idea-generating group, although it is realized in a rather simple form, which involves the individual memory as well as sharing, (pre- and post-) idea selection, and idea generation mechanisms. 

First, each agent possesses an individual memory that can only be modified by the agent itself. For an agent, memory \cite{Glenberg:1997p1390} is essential for supporting its individual learning capability in utilizing its past experience. Here the memory contains only one base idea $\vec{x}_{A}$.

Second, each agent shares (parts of) its memory as nonprivate information \cite{Danchin:2004p1204}, as a basic sharing mechanism to affect the group \cite{Liu:2006p1045}. Here the whole $\vec{x}_{A}$ is contributed. 

Third, each agent has the capability to utilize ideas that are contributed by others \cite{Paulus:2000p1114}. For real-world animals, the social learning capability \cite{Galef:1995p1128,MontesdeOca2011} can enhance their adaptability in a changing environment, which may lead to a cumulative evolution of ideas that are more novel and useful than those contributed by individuals \cite{Kohn2011}. In each session, each agent only accepts one social idea, called $\vec{x}_{S}$, from the group memory in its environment. 

Finally, the ideation capability of each agent is accomplished by a socially-biased combine-and-improve procedure, as shown in Figure \ref{Fig:RTGO}(b). The pre-selection mechanism simply returns $\vec{x}_{A}$ and $\vec{x}_{S}$. The idea combination process forms a new idea $\vec{x}_{C}$ by combining $\vec{x}_{A}$ and $\vec{x}_{S}$. Afterward, the local improvement process is applied for further improving $\vec{x}_{C}$. The post-selection mechanism is then used to update the individual memory using $\vec{x}_{C}$.

Algorithm \ref{alg:Anticipated} gives the working process of RTGO. Initially, each agent owns an idea that is randomly generated in the problem space (Line 1). Then RTGO runs in iterative sessions. Lines 3-6 describe the simple round-table mechanism that provides $\vec{x}_{S}$ for each agent. In Line 8, each agent runs the combine-and-improve procedure, including the idea combination process and the local improvement process, which are respectively realized using a XS strategy an a LS strategy, to generate a promising idea $\vec{x}_{C}$. Under the RTGO framework, the combine-and-improve procedure can be realized in the form of socially-biased learning \cite{Galef:1995p1128} that using both the base and social ideas, where the base idea serves as an incumbent solution, and the social idea provides stimulation clues in different sessions. For each agent, $\vec{x}_{A}$ is a steady-state idea over sessions since it is replaced by $\vec{x}_{C}$ only if  $\vec{x}_{C}$ has a better quality. Finally, the best idea $\vec{x}^{*}$  of all agents is the solution of group brainstorming.

\begin{algorithm} [t]                     
\caption{Round-table group optimizer (RTGO)}   
\label{alg:Anticipated}                           
\begin{algorithmic}[1]                    
\STATE Initially, each agent owns an idea $\vec{x}_{A}$ that is randomly generated in the problem space 

\FOR{Session $t = 1$ to $T$}
\STATE The agents randomly take the seats (labelled from 1 to $N$) around a round table
\FOR{Agent $h = 1$ to $N$}
\STATE $\vec{x}_{S,((h+1) \operatorname{mod} N)}$ = $\vec{x}_{A,h}$~~~~~~~~~\COMMENT{Agent $h$ shares its base idea to its clockwise neighbor}
\ENDFOR
\FOR{Agent $h = 1$ to $N$}
\STATE $\vec{x}_{C,h} = \text{XS}(\vec{x}_{A,h}, \vec{x}_{S,h})$; $\vec{x}_{C,h} = \text{LS}(\vec{x}_{C,h})$~~~~~~~~\COMMENT{socially-biased combine-and-improve}
\STATE {\bf if} $f(\vec{x}_{A,h}) \geq f(\vec{x}_{C,h})$ {\bf then} $\vec{x}_{A,h} = \vec{x}_{C,h}$~~~~~~~~~~~~~~~~~~~~~~~~~~~~~~~\COMMENT{steady-state memory}
\ENDFOR
\ENDFOR
\RETURN{$\vec{x}^{*}  = (\vec{x}_{A,h}$ with the minimum objective value for $h\in[1, N]$})
\end{algorithmic}
\end{algorithm}

RTGO has two overall setting parameters, i.e., the number of agents $N$ and the number of sessions $T$, and XS and LS strategies are to be implemented for specific problems.

\section{Implementation on Sequencing Problems}
\label{sec:Permutation}
Sequencing problems have the same problem space. For a problem with $n$ nodes, each state $\vec{x}$ is an array containing a permutation \{$x_{1}$, \dots , $x_{n}$\} of the integer value from 1 through $n$. However, different sequencing landscapes might differ in domain-specific structures.

\subsection{Problem Description}

In this paper, we consider two important examples of sequencing problems, i.e., the flowshop scheduling problem (FSP) and the quadratic assignment problem (QAP).

\subsubsection{Flowshop Scheduling Problem (FSP)}

The flowshop scheduling problem (FSP) consists in scheduling $n$ independent jobs to be processed on $m$ independent machines in the same order. At any time, each job has one operation on one machine and each machine can process only one job. There is an $n \times m$ matrix of processing times $P$=($p_{ij}$), where each $p_{ij}$ is the processing time of the operation of the $i$th job on the $j$th machine. Each state $\vec{x}$, i.e., a schedule, is a job processing permutation, where $x_{i} (i \in [1, n])$ denotes the processing order of the $i$th job on every machine.

For each schedule $\vec{x}$, the completion time $c(\vec{x}, i, j)$ of the $i$th job on the $j$th machine is 

\begin{equation}
c(\vec{x}, i, j) = \operatorname{max} (c(\vec{x}, i-1, j), c(\vec{x}, i, j-1))+p_{x_{i} j},
\end{equation}
where $c(\vec{x}, i, 0) = 0$ for $\forall i\in [1, n]$ and $c(\vec{x}, 0, j) = 0$ for $\forall j\in [1, m]$.

The objective function is the total completion time (or makespan) of each $\vec{x}$, i.e.,

\begin{equation}
f(\vec{x}) = c(\vec{x}, n, m).
\end{equation}

\subsubsection{Quadratic Assignment Problem (QAP)}

The quadratic assignment problem (QAP) consists of assigning $n$ facilities to $n$ locations, one facility at a location. There are two $n \times n$ matrices, i.e., the flow matrix $W$ = ($w_{ij}$) and the distance matrix $D$=($d_{ij}$), in which $w_{ij}$ is the flow between facilities $i$ and $j$, and $d_{ij}$ is the distance between locations $i$ and $j$. The objective function is 

\begin{equation}
f(\vec{x}) =\sum _{i=1}^{n}\sum _{j=1}^{n}d_{ij} w_{x_{i}x_{j}}.
\label{eqn:QAP}
\end{equation}

In addition, an asymmetric instance is preprocessed into a symmetric one without changing the resulting cost, if one of the two matrices $D$ and $W$ is symmetric \cite{Merz:2000p2516}.

\subsection{Local Improvement}

For the local improvement process, two stable LS strategies are respectively used for the two sequencing problems. Each local search strategy starts from an incumbent state $\vec{x}$ and then tries to improve it by executing basic neighborhood moves in a systematic way. 

For FSP, we consider an insertion-based local search operator \cite{Stutzle:1998p2467, Ruiz:2006p2435, Ruiz:2007p2496}, as shown in Algorithm \ref{alg:LS_FSP}. The basic operation $BestInsert(\vec{x},k_i)$ returns the best permutation obtained by inserting a job $k_i$ to any possible position of $\vec{x}$. Although each objective function alone needs to be calculated in the time complexity $O(n\cdot m)$, the total $(n-1)$ positions can be checked in $O(n\cdot m)$ as well, based on the speed-up technique proposed by Taillard \cite{Taillard:1990p2487}.

\begin{algorithm} [h]                     
\caption{Iterative insertion-based local search for each incumbent state $\vec{x}$ of FSP}   
\label{alg:LS_FSP}                           
\begin{algorithmic}[1]                    
\STATE $isImproved=$ true; $f_{O}=f(\vec{x})$
\WHILE{$isImproved$}
\STATE $isImproved=$ false; Generate a random sequence $({k_1, \cdots, k_n})$ for $k_i \in[1,n]$
\FOR{$i = 1$ to $n$} 
\STATE  $\vec{x} = BestInsert(\vec{x},k_i)$
\IF {$f(\vec{x}) < f_{O}$}
\STATE  $f_{O} = f(\vec{x})$; $isImproved=$ true
\ENDIF
\ENDFOR
\ENDWHILE
\end{algorithmic}
\end{algorithm}

For QAP, we consider an exchange-based LS operator \cite{Gambardella:1999p2521, Ramkumar:2008p2501}, which runs twice of the complete neighborhood search in Algorithm \ref{alg:LS_QAP}. It differs from \textit{fast-}2\textit{-opt} \cite{Merz:2000p2516} only in that the latter one iteratively runs Algorithm \ref{alg:LS_QAP} until no improvement can be found.

\begin{algorithm} [h]                     
\caption{Exchange-based neighborhood search for each incumbent state $\vec{x}$ of QAP}   
\label{alg:LS_QAP}                           
\begin{algorithmic}[1]                    
\STATE Generate two random sequences $({r_1, \cdots, r_n})$ and $({s_1, \cdots, s_n})$ for $r_i \in[1,n]$ and $s_j \in[1,n]$
\FOR{$i = 1$ to $n$, $j = 1$ to $n$} 
\STATE {\bf if} {$\Delta Exchange(\vec{x},r_i,s_j) < 0$} {\bf then} Exchange $r_i$ and $s_j$ in $\vec{x}$
\ENDFOR
\end{algorithmic}
\end{algorithm}

The delta quality of each exchange between locations $i$ and $j$ can be calculated as \cite{Taillard:1991p2451}:
\begin{equation}
\begin{array}{l} {\Delta Exchange(\vec{x},i,j)=(d_{ii} -d_{jj} )\cdot (w_{x_{j} x_{j} } -w_{x_{i} x_{i} } )+(d_{ij} -d_{ji} )\cdot (w_{x_{j} x_{i} } -w_{x_{i} x_{j} } )} \\ 
{{\rm \; \; \; \; \; }+\sum _{k=1,k\ne i,j}^{n}((d_{ki} -d_{kj} )\cdot (w_{x_{k} x_{j} } -w_{x_{k} x_{i} } )+(d_{ik} -d_{jk} )\cdot (w_{x_{j} x_{k} } -w_{x_{i} x_{k} } )) } \end{array}
\end{equation}

If both matrices $W$ and $D$ are symmetric, and that all diagonal elements of either matrix are zeros, the delta evaluation can be simplified as \cite{Taillard:1991p2451}:
\begin{equation}
\begin{array}{l} {\Delta Exchange(\vec{x},i,j)=2\cdot \sum _{k=1,k\ne i,j}^{n}(d_{ik} -d_{jk} )(w_{x_{j} x_{k}}-w_{x_{i}x_{k}} ) }\end{array}
\end{equation}

Both delta evaluation methods are computable in \textit{O}($n$), and the latter one is a bit faster than the former one. In addition, an asymmetric instance can be converted into a symmetric one if one of the two matrices is symmetric \cite{Merz:2000p2516}.

\subsection{Idea Combination}

The idea combination is realized by a XS strategy, which produces one output idea, i.e., $\vec{x}_{C}$, by using information from two input ideas, i.e., the base idea $\vec{x}_{A}$ and the social idea $\vec{x}_{S}$.

We define a class of XS strategies, called {\em socially biased combination} (SBX), where $\vec{x}_{A}$ and $\vec{x}_{S}$ are viewed as the incumbent idea and social guidance information, respectively. 

SBX operators can be realized in basic or macro forms. Each basic SBX operator contains three policies, i.e., a \textit{base policy}, a \textit{social policy}, and a \textit{repair policy}. Initially, all positions in $\vec{x}_{C}$ are unoccupied. Then the three policies are sequentially executed to occupy positions in $\vec{x}_{C}$ with unused values, so that all positions are eventually occupied. Each macro SBX operator can be realized by applying a {\em macro policy} on basic and macro SBX operators.

\subsubsection{Base Policy}

The base policy marks the nodes from certain positions of the base state $\vec{x}_{A}$, and then copies the values at the marked nodes into the corresponding positions of $\vec{x}_{C}$. 

There are three commonly-used modes. In the \textit{one-point} mode (1P) \cite{Murata:1996p2463}, one cut-point is randomly selected, and then the set of nodes on a randomly chosen side of the cut-point are marked. In the \textit{two-point} mode (2P) \cite{Murata:1996p2463}, a pair of cut-points is randomly selected, then the nodes located on either inside or outside of the selected two cut-points are marked. In the \textit{uniform} mode (U) \cite{Syswerda:1991p2518}, each position is marked with the probability of 0.5. All these modes preserve position-based information. The difference is in that 1P, 2P and U provide \textit{O}($n$), \textit{O}($n^{2}$), and \textit{O}($2^{n}$) numbers of choices, respectively. Furthermore, the 1P mode preserves whereas the U mode loses most precedence and adjacency information in the base idea.

Furthermore, a \textit{common-avoiding} one-point mode (CA1P) is proposed. CA1P differs from 1P only in that the common nodes from the left and right sides are not considered when randomly selecting the cut-point. For example, for two states \{\textit{2 6 1} 3 4 9 5 8 \textit{7}\} and \{\textit{2 6} \textit{1} 4 8 5 3 9 \textit{7}\}, the first three nodes and the last one node are the common nodes at both sides, and thus the cut-point is randomly chosen between the 4th location and the 7th location. CA1P ensures the marked nodes do not contain fully common values.

\subsubsection{Social Policy}

The social policy fills some remaining unoccupied positions of $\vec{x}_{C}$ by using unused nodes at certain selected positions of the social state $\vec{x}_{S}$. In the \textit{position-based} mode (P), the unused values of the state $\vec{x}_{S}$ are copied to corresponding positions of $\vec{x}_{C}$ if such positions are unoccupied. Hence, this mode preserves position and precedence information of the selected nodes of $\vec{x}_{S}$. In the \textit{order-based} mode (O), all unused nodes of $\vec{x}_{S}$ are respectively copied to the unoccupied positions from left to right. Hence, this mode preserves precedence information of the selected nodes of $\vec{x}_{S}$. Furthermore, the order-based mode can always achieve a valid state, wheres the position-based mode may leave some positions unoccupied.

\subsubsection{Repair Policy}

The repair policy turns $\vec{x}_{C}$ into a valid state if there are any unoccupied positions. Here two modes are realized. The \textit{random} mode (R) fills each unoccupied position by a randomly selected unused value. The \textit{partially mapped} mode (PM), i.e., the final step in the partially mapped crossover \cite{Starkweather:1991p2505}, uses the position-based mappings between the nodes of both parent states to fill remaining positions for further preserving the order and position.

\subsubsection{Macro Policy}

In this paper, we only considered a simple macro policy, called the \textit{parallel} mode (MP), which outputs the state with the best quality among the valid candidate states generated by totally $N_{ICG}$ independent trials of a component XS operator. 

The parallel mode can be viewed as an inner portfolio of algorithms \cite{Gomes:2001p2488}, since the component XS operator can be seemed as a stochastic algorithm. An algorithm portfolio may avoid the heavy tails of individual trials, thus the output state can be more promising.

\subsubsection{Summary}

Each SBX case can be formally described by specifying its components, so that it is easy to know the subtle similarity and difference between various strategies. 

A basic SBX case is denoted as a combination of three tags, i.e., A/B/C, where the tags A, B, and C designates the modes of base, social, and repair policies, respectively. Moreover, a tag is denoted as ``-'' if the corresponding policy is not required, or as ``*'' if the corresponding policy has not been assigned. If the parallel macro policy is applied on a basic SBX case A/B/C, then the macro SBX case is defined as MP(A/B/C).

Some existing XS operators can be approximately represented as SBX cases. One-point \cite{Murata:1996p2463}, two-point, and uniform (or position-based crossover \cite{Starkweather:1991p2505}) order crossovers, are respectively similar to 1P/O/-, 2P/O/-, and U/O/-. LCS crossover \cite{Iyer:2004p2489} and similar block crossovers \cite{Ruiz:2006p2435} differ from these simple order crossovers only in that they used more delicate base policies. The partially mapped crossover \cite{Starkweather:1991p2505} and its uniform variant \cite{Cicirello:2000p2500} are similar to 2P/P/PM and U/P/PM, respectively. The uniform like crossover (ULX) \cite{Tate:1995p2433} and its optimized variant \cite{Misevicius:2004p2519}  are similar to U/P/R and MP(U/P/R), respectively.

\section{Results and Discussion}
\label{sec:Comparisons}

All RTGO versions are coded in JAVA, and were run on a 1.4-GHz Opteron processor. By default, the maximum numbers of cycles ($T$) are fixed as 100 and 500 for FSP and QAP, respectively. For the parallel macro mode, there is $N_{ICG} ={\rm MAX}({\rm INT}(0.05 \cdot n),{\rm \; }1)$, in which INT returns the closest integer value, MAX returns the larger value.

For FSP, experiments were performed on four benchmark datasets in OR-Library\footnote{http://people.brunel.ac.uk/$\sim$mastjjb/jeb/orlib/flowshopinfo.html} \cite{Beasley:1990p2481}, namely, Carlier's, Heller's , Reeves's, and Taillard's datasets. The Carlier's dataset contains 8 instances named Car1, Car2, through Car8. The Heller's dataset only possesses 2 instances, i.e., Hel1 and Hel2. The Reeves's dataset has 21 instances called Rec01, Rec03, through Rec41. The Taillard's dataset contains 90 instances from Tai01 through Tai90. 

Table 1 lists the best-known upper bounds (\textit{f}*) of the Taillard's dataset lasted updated on 13-APR-2005 (or called as UB05)\footnote{http://mistic.heig-vd.ch/taillard/problemes.dir/ordonnancement.dir/flowshop.dir/best\_lb\_up.txt}. Besides, some researchers used the upper bounds reported by Taillard in 1993 (UB93) \cite{Taillard:1993p2472}, or the upper bounds listed in 2004 (UB04) \cite{Ruiz:2006p2435}, etc. Basically, UB05 is much better than UB93, and slightly better than UB04.

\begin{table} [t]
\centering \caption{The best-known upper bounds of the Taillard's dataset on 13-APR-2005 (UB05).}
\begin{tabular}{|c|c|c|c|c|c|c|c|c|c|} \hline 
No. & 01$\sim$10 & 11$\sim$20 & 21$\sim$30 & 31$\sim$40 & 41$\sim$50 & 51$\sim$60 & 61$\sim$70 & 71$\sim$80 & 81$\sim$90 \\ \hline 
1 & 1278 & 1582 & 2297 & 2724 & 2991 & 3847 & 5493 & 5770 & 6202 \\ \hline 
2 & 1359 & 1659 & 2099 & 2834 & 2867 & 3704 & 5268 & 5349 & 6183 \\ \hline 
3 & 1081 & 1496 & 2326 & 2621 & 2839 & 3640 & 5175 & 5676 & 6271 \\ \hline 
4 & 1293 & 1377 & 2223 & 2751 & 3063 & 3719 & 5014 & 5781 & 6269 \\ \hline 
5 & 1235 & 1419 & 2291 & 2863 & 2976 & 3610 & 5250 & 5467 & 6314 \\ \hline 
6 & 1195 & 1397 & 2226 & 2829 & 3006 & 3679 & 5135 & 5303 & 6364 \\ \hline 
7 & 1234 & 1484 & 2273 & 2725 & 3093 & 3704 & 5246 & 5595 & 6268 \\ \hline 
8 & 1206 & 1538 & 2200 & 2683 & 3037 & 3691 & 5094 & 5617 & 6401 \\ \hline 
9 & 1230 & 1593 & 2237 & 2552 & 2897 & 3741 & 5448 & 5871 & 6275 \\ \hline 
10 & 1108 & 1591 & 2178 & 2782 & 3065 & 3756 & 5322 & 5845 & 6434 \\ \hline 
\end{tabular}
\label{Tab:TalDataSet}
\end{table}

For QAP, fifty commonly-used instances from QAPLIB \cite{Burkard:1997p2507} are used. The names and best-known upper bounds of the instances will be listed in the tables in Section \ref{sec:QAP}.

There are two significant indices for measuring the performance of an algorithm. The first is the solution quality, which can be represented by relative percentage deviation (RPD) over the best-known upper bound (\textit{f}${}^{*}$). The second is running time ($t_{r}$) in seconds, which is counted at the cycle taken to reach the last improvement. Only $t_{r}$ might be influenced by different machine configurations for running an algorithm. For the FSP and QAP instances, 10 and 50 independent runs were executed to obtain the mean results, respectively.


\subsection{Effects of Idea Combination}
\label{sec:IdeaComb}

During the search process, the idea combination process plays a navigating role for exploring the rugged permutation landscape by guiding the greedy local improvement process. 

For comparison the differences between RTGO with different idea combination operators, RPD-$t_{r}$ relations are used for examining the Pareto efficiency of both performance indices.

\subsubsection{For FSP}

For FSP, the effects of XS operators are evaluated on the Taillard's dataset. Moreover, 1P/O/-, i.e., the one-point order crossover \cite{Murata:1996p2463}, is considered as a standard XS operator for the comparisons. For each case, RTGO with $N$=10, 20, 30, 40, and 50 were tested.

Three order-based operators (*/O/-), i.e., 1P/O/-, 2P/O/-, and U/O/-, have been frequently studied, where the difference is that they using different base policies from the viewpoint of SBX. Some researchers \cite{Murata:1996p2463,Nearchou:2004p2452} have claimed that 1P/O/- performs better than both 2P/O/- and U/O/-, However, some other researchers have declared \cite{Iyer:2004p2489} that 1P/O/- is the best among the above versions, and some advanced versions, such as the similar block order crossover (SBOX) \cite{Ruiz:2006p2435}, are modified from 1P/O/-. 

\begin{figure} [p]
\centering \includegraphics*[width=2.85in]{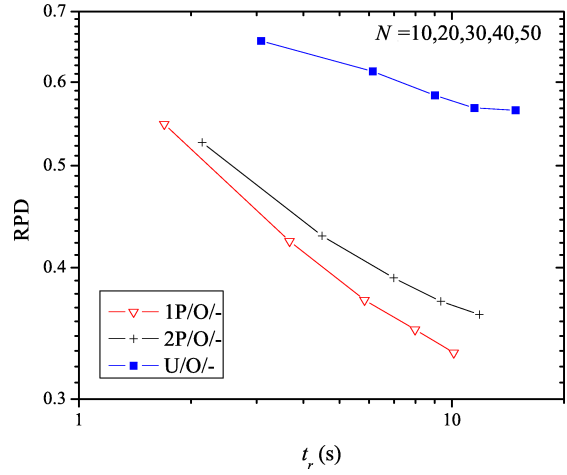}
\caption{Results of the RTGO versions with SBX in different base policies.}
\label{Fig:fig1}
\end{figure}

\begin{figure} [p]
\centering \includegraphics*[width=2.85in]{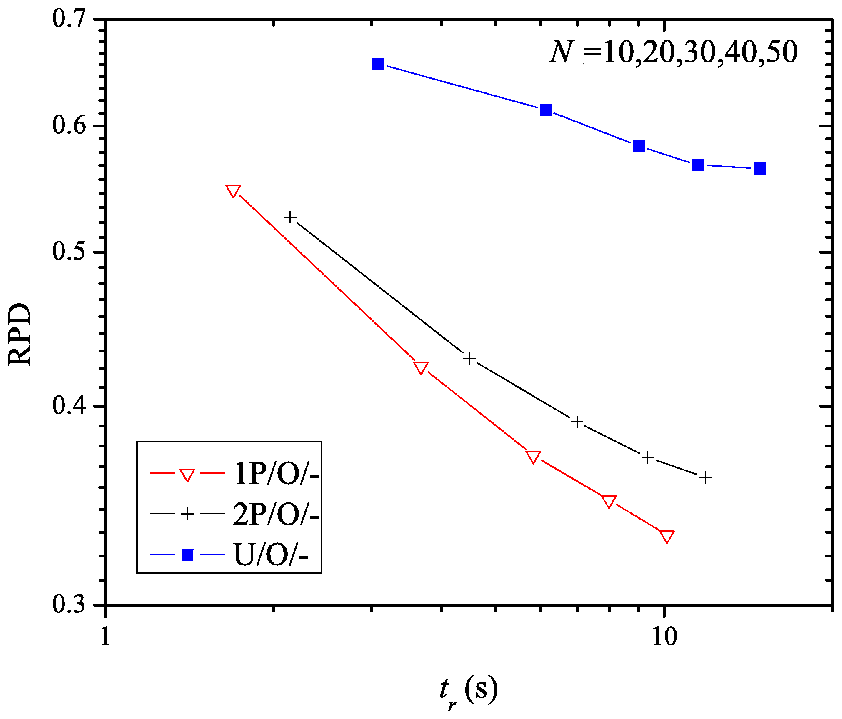}
\caption{Results of the RTGO versions with SBX in different social and repair policies.}
\label{Fig:fig2}
\end{figure}

\begin{figure} [p]
\centering \includegraphics*[width=2.85in]{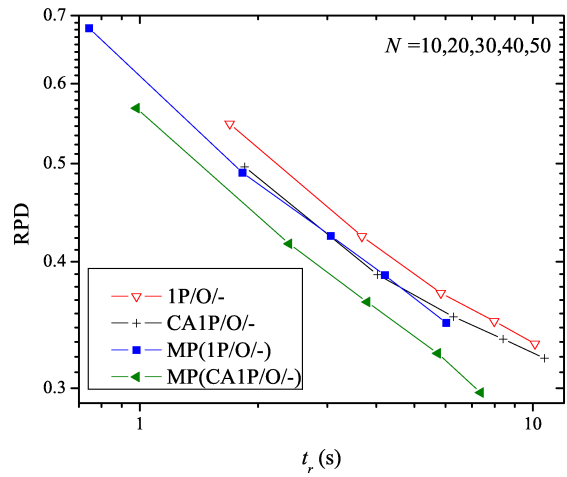}
\caption{Results of the RTGO versions with some improving heuristics.}
\label{Fig:fig3}
\end{figure}

Figure \ref{Fig:fig1} gives the results by the RTGO versions using 1P/O/-, 2P/O/-, and U/O/-, respectively. If only RPD is concerned, contradicting conclusions may be achieved as for RTGO using different $N$ values, where 2P/O/- has a lower RPD value than 1P/O/- as $N$=10, but has higher RPD values than 1P/O/- as $N$=20, 30, 40, and 50. However, it can be found that 1P/O/- has an overall better RPD-$t_{r}$ performance than 2P/O/-. Moreover, U/O/- has the worst performance among all the three SBX operators, which may due to that the uniform mode preserves much less precedence information.

Figure \ref{Fig:fig2} provides the results by the RTGO versions using 1P/O/-, 1P/P/O, and 1P/P/R, respectively. Here 1P/O/- performs better than the other two operators. Compared with 1P/P/*, 1P/O/- preserves more precedence information. For the repair operation, using the order-based mode is slightly better than using the random node. Thus for FSP, precedence information may be more significant than position information. 

Figure \ref{Fig:fig3} shows the results by the RTGO versions using 1P/O/-, CA1P/O/-, MP(1P/O/-) and MP(CA1P/O/-), respectively. Both CA1P/O/- and MP(1P/O/-) obtain better RPD-$t_{r}$ performances than 1P/O/-. Specifically, CA1P/O/- achieves a much better RPD although it needs a longer running time, since the common-avoiding mode maintains larger information diversity among the agents, especially as $N$ is smaller; while MP(1P/O/-) achieves a much shorter running time although it obtains a worse RPD, since the macro mode in the parallel mode leads to more promising states for the LS operator. MP(CA1P/O/-), which integrates of both improving strategies, produces better RPD-$t_{r}$ performance than all three others.

\subsubsection{For QAP}

\begin{figure} [p]
\centering \includegraphics*[width=2.85in]{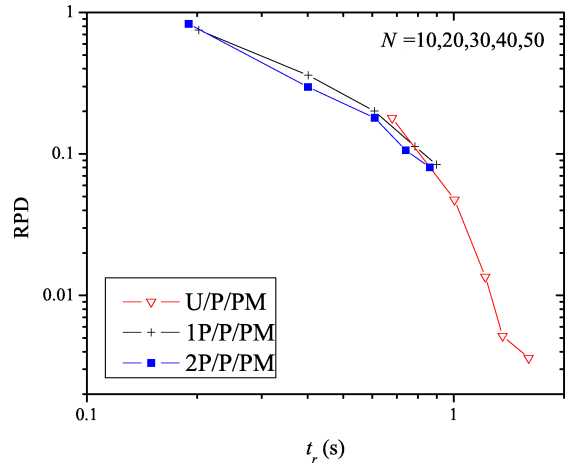}
\caption{Results of the RTGO versions with SBX in different base policies.}
\label{Fig:fig4}
\end{figure}

\begin{figure} [p]
\centering \includegraphics*[width=2.85in]{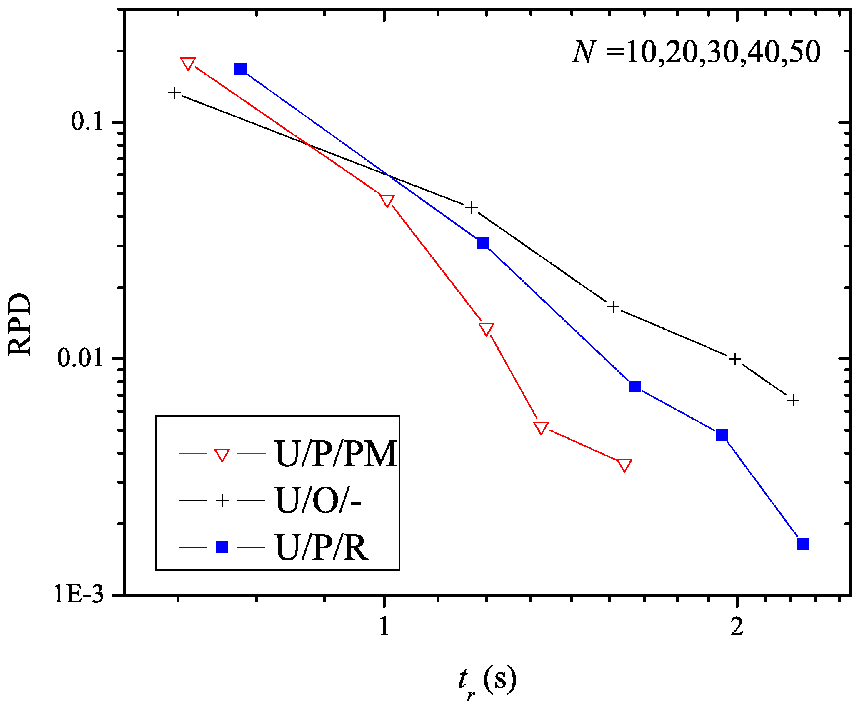}
\caption{Results of the RTGO versions with SBX in different social and repair policies.}
\label{Fig:fig5}
\end{figure}

\begin{figure} [p]
\centering \includegraphics*[width=2.85in]{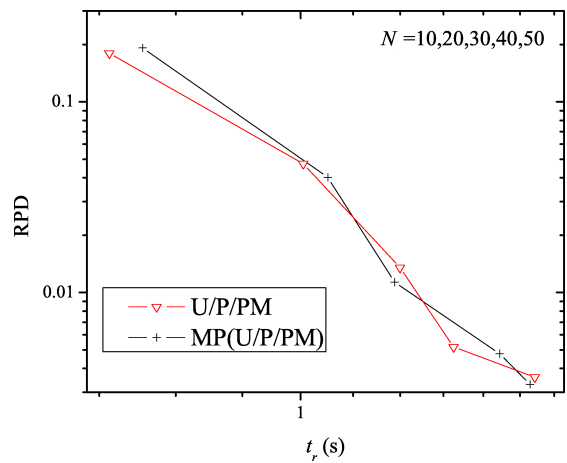}
\caption{Results of the RTGO versions with U/P/PM and MP(UP/P/PM).}
\label{Fig:fig6}
\end{figure}

For QAP, the effects of XS operators are evaluated on the 50 instances in QAPLIB. Moreover, U/P/PM, i.e., the uniform partially mapped crossover \cite{Cicirello:2000p2500}, which is previously used for solving the largest common subgraph problem, is used as a standard XS operator for the comparisons. For each case, RTGO with $N$=10, 20, 30, 40, and 50 were tested.

Figure \ref{Fig:fig4} gives the results by the RTGO versions respectively using U/P/PM, 1P/P/PM, and 2P/P/PM. As $N$ is same, U/P/PM can achieve a lower RPD yet require a larger $t_{r}$ than both 1P/P/PM and 2P/P/PM, which may due to that the U mode provides more possible choices than the 1P or 2P mode. Moreover, the base policy in 1P or 2P mode did not show an advantage in terms of the RPD-$t_{r}$ relation.

Figure \ref{Fig:fig5} shows the results by the RTGO versions using U/P/PM, U/O/-, and U/P/R, respectively. Both UP/P/R and UP/P/PM perform better than U/O/-. Here U/P/R represents the uniform like crossover (ULX) \cite{Tate:1995p2433}. Compared with U/O/-, U/P/* preserves more position-based information. For the repair operation, It showed that using the PM mode is better than using the R mode, since the PM mode also preserves more position information. Thus for QAP, the position information might be more significant.

Figure \ref{Fig:fig6} shows the results by the RTGO versions respectively using UP/P/PM and MP(UP/P/PM). Their performance are similar from the RPD-$t_{r}$ relation.

\subsection{Comparisons with Existing Algorithms}
\label{sec:ExistingAlgo}

RTGO is coded in JAVA and most algorithms are coded in C/C++. Since the performance gap between JAVA and C++ is close, it only requires comparing the running times of different algorithms on different hardware configurations. 

For simplicity, we calculate $r_{TR} =(t_{r,{\rm A}} /t_{r,{\rm B}} {\rm )}\cdot {\rm (}CR_{{\rm A}} /CR_{{\rm B}} )$, in which $t_{r,{\rm A}} $ and $t_{r,{\rm B}} $ are the running times of the algorithms A and B, \textit{CR}$_{A}$ and \textit{CR}$_{B}$ are the clock rates of the processors for running the algorithms A and B, respectively. Moreover, the ratio of CPU clock rates is eliminated if no clock rate is provided for any of the algorithms. Here A and B are RTGO and the algorithm to be compared, respectively. In addition, a threshold value $r_{{\rm MAX}} $ is used for taking the factors of other configurations into account. For $r_{TR} \le 1/r_{{\rm MAX}} $, $1/r_{{\rm MAX}} <r_{TR} <r_{{\rm MAX}} $, and $r_{TR} \ge r_{{\rm MAX}} $, RTGO are slower than, comparable to, and faster than the compared algorithm, respectively. In this paper, $r_{{\rm MAX}} $=5 is used.

\subsubsection{For FSP}

In the following experiments, only the RTGO using MP(CA1P/O/-) is considered for comparing with some existing algorithms on the FSP benchmark datasets.

\begin{table} [b]
\centering \caption{Results by RTGO, PSOMA, and HW-LS on the Carlier's dataset.}
\begin{tabular}{|c|c|c|c|c|c|c|c|c|} \hline 
\multirow{2}{*}{Instance} & \multirow{2}{*}{$n$, $m$} & \multirow{2}{*}{\textit{f}${}^{*}$} & \multicolumn{2}{|c|}{RTGO($N$=30)} & \multicolumn{2}{|c|}{PSOMA} & \multicolumn{2}{|c|}{HW-LS} \\ \cline{4-9}
 &  &  & RPD & $t_{r}$(s) & RPD & $t_{r}$(s) & RPD & $t_{r}$(s) \\ \hline 
Car1 & 11,5 & 7038 & 0.000  & 0.008 & 0.000  & 0.68  & 0.00  & 0.010  \\ \hline 
Car2 & 13,4 & 7166 & 0.000  & 0.009 & 0.000  & 0.95  & 0.00  & 0.020  \\ \hline 
Car3 & 12,5 & 7312 & 0.000  & 0.015 & 0.000  & 1.06  & 0.00  & 0.020  \\ \hline 
Car4 & 14,4 & 8003 & 0.000  & 0.009 & 0.000  & 1.22  & 0.00  & 0.010  \\ \hline 
Car5 & 10,6 & 7720 & 0.000  & 0.011 & 0.018  & 0.70  & 0.00  & 0.020  \\ \hline 
Car6 & 8,9 & 8505 & 0.000  & 0.007 & 0.114  & 0.49  & 0.00  & 0.010  \\ \hline 
Car7 & 7,7 & 6590 & 0.000  & 0.006 & 0.000  & 0.30  & 0.00  & 0.010  \\ \hline 
Car8 & 8,8 & 8366 & 0.000  & 0.007 & 0.000  & 0.42  & 0.00  & 0.010  \\ \hline 
Average & - & - & 0.000  & 0.009 & 0.017  & 0.73 & 0.00  & 0.014  \\ \hline 
\end{tabular}
\label{Tab:CarlierSet}
\end{table}

\begin{table} [t]
\centering \caption{Results by RTGO, PSOMA, and HW-LS on the Reeves's dataset.}
\begin{tabular}{|c|c|c|c|c|c|c|c|c|} \hline 
\multirow{2}{*}{Instance} & \multirow{2}{*}{$n$, $m$} & \multirow{2}{*}{\textit{f}${}^{*}$} & \multicolumn{2}{|c|}{RTGO($N$=30)} & \multicolumn{2}{|c|}{PSOMA } & \multicolumn{2}{|c|}{HW-LS} \\ \cline{4-9}
 &  &  & RPD & $t_{r}$(s) & RPD & $t_{r}$(s) & RPD & $t_{r}$(s) \\ \hline 
Rec01 & 20,5 & 1247 & 0.096  & 0.040  & 0.144  & 2.60  & 0.02  & 5.57 \\ \hline 
Rec03 & 20,5 & 1109 & 0.000  & 0.048  & 0.189  & 2.50  & 0.00  & 2.57 \\ \hline 
Rec05 & 20,5 & 1242 & 0.242  & 0.016  & 0.249  & 2.39  & 0.24  & 10.4 \\ \hline 
Rec07 & 20,10 & 1566 & 0.000  & 0.104  & 0.986  & 2.81  & 0.00  & 1.86 \\ \hline 
Rec09 & 20,10 & 1537 & 0.000  & 0.093  & 0.621  & 4.23  & 0.00  & 2.64 \\ \hline 
Rec11 & 20,10 & 1431 & 0.000  & 0.065  & 0.129  & 3.79  & 0.00  & 0.77 \\ \hline 
Rec13 & 20,15 & 1930 & 0.109  & 0.191  & 0.893  & 4.64  & 0.09  & 46.4 \\ \hline 
Rec15 & 20,15 & 1950 & 0.021  & 0.274  & 0.628  & 5.23  & 0.44  & 35.3 \\ \hline 
Rec17 & 20,15 & 1902 & 0.037  & 0.225  & 1.330  & 4.67  & 0.11  & 24.2 \\ \hline 
Rec19 & 30,10 & 2093 & 0.287  & 0.545  & 1.313  & 10.49  & 0.63  & 82.8 \\ \hline 
Rec21 & 30,10 & 2017 & 1.120  & 0.435  & 1.596  & 8.41  & 1.41  & 74.2 \\ \hline 
Rec23 & 30,10 & 2011 & 0.373  & 0.586  & 1.310  & 9.36  & 0.54  & 81.0 \\ \hline 
Rec25 & 30,15 & 2513 & 0.294  & 0.765  & 2.085  & 12.64  & 1.11  & 127 \\ \hline 
Rec27 & 30,15 & 2373 & 0.324  & 0.794  & 1.605  & 12.15  & 0.94  & 137 \\ \hline 
Rec29 & 30,15 & 2287 & 0.249  & 0.815  & 1.888  & 11.31  & 0.80  & 141 \\ \hline 
Rec31 & 50,10 & 3045 & 0.263  & 2.005  & 2.254  & 37.15  & 1.91  & 292 \\ \hline 
Rec33 & 50,10 & 3114 & 0.167  & 0.979  & 0.645  & 36.07  & 0.47  & 228 \\ \hline 
Rec35 & 50,10 & 3277 & 0.000  & 0.182  & 0.000  & 29.92  & 0.00  & 6.40 \\ \hline 
Rec37 & 75,20 & 4951 & 0.913  & 14.095  & 3.547  & 170.2  & 4.19  & 1700 \\ \hline 
Rec39 & 75,20 & 5087 & 0.698  & 11.966  & 2.426  & 155.7  & 2.90  & 1680 \\ \hline 
Rec41 & 75,20 & 4960 & 1.183  & 12.859  & 3.684  & 164.3  & 3.93  & 1910 \\ \hline 
Average & - & - & 0.304  & 2.242  & 1.311  & 32.88  & 0.94  & 313.8 \\ \hline 
\end{tabular}
\label{Tab:ReevesSet}
\end{table}
Tables \ref{Tab:CarlierSet} and \ref{Tab:ReevesSet} give the results by the RTGO with $N$=30, PSOMA \cite{Liu:2007p2457}, and HW-LS \cite{Huang:2006p2522} on the Carlier's and Reeves's datasets, respectively. PSOMA \cite{Liu:2007p2457} is a particle swarm optimization (PSO)-based memetic algorithm, which is coded in MATLAB 7.0, and was executed on a 2.2-GHz Mobile Pentium IV processor. HW-LS \cite{Huang:2006p2522} is a LS method combined with two escape-from-trap procedures, which was run on a 500-MHz Pentium III processor.

Table \ref{Tab:CarlierSet} indicates that RTGO performed better than PSOMA for both RPD and running time on the Carlier's dataset. Both RTGO and HW-LS solved all instances in a 100\% success rate, i.e., they are efficient in moving away from local minima.

From Table \ref{Tab:ReevesSet}, it can be found that RTGO produced much better than both PSOMA and HW-LS for both RPD and running time on the Reeves's dataset. The $r_{TR} $ values of RTGO versus PSOMA and HW-LS are 127.5 and 0.6 for the Carlier's dataset, 23.0 and 50.0 for the Reeves's dataset, respectively. If considering the machine configurations, RTGO was not faster than HW-LS on the Carlier's dataset. However, RTGO was much faster than HW-LS on the Reeves's dataset. It may due to the Carlier's instances are easier than the Reeves's.

\begin{table} [htbp]
\centering \caption{ Results by RTGO, DPSO2, and NEH-ALA on the Heller's dataset.}
\begin{tabular}{|c|c|c|c|c|c|c|c|c|} \hline 
\multirow{2}{*}{Instance} & \multirow{2}{*}{$n$, $m$} & \multirow{2}{*}{\textit{f}${}^{*}$} & \multicolumn{2}{|c|}{RTGO($N$=30)} & \multicolumn{2}{|c|}{PSOMA } & \multicolumn{2}{|c|}{HW-LS} \\ \cline{4-9}
 &  &  & RPD & $t_{r}$(s) & RPD & $t_{r}$(s) & RPD & $t_{r}$(s) \\ \hline 
Hel1 & 100,10 & 514 & 0.156 & 1.412 & 1.15 & 287 & 3.55 & 94 \\ \hline 
Hel2 & 20,10 & 135 & 0.000 & 0.153 & 1.08 & 67 & 0.39 & 2565 \\ \hline 
\end{tabular}
\label{Tab:HellerSet}
\end{table}

\begin{table} [t]
\centering \caption{Results by some existing algorithms on the Taillard's dataset.}
\begin{tabular}{|c|c|c|c|c|c|c|c|c|} \hline 
\multirow{2}{*}{Instance} & HSA & PACO & \multicolumn{2}{|c|}{ILS} & \multicolumn{2}{|c|}{HGA\_RMA} & \multicolumn{2}{|c|}{PSO$_{VNS}$} \\ \cline{2-9} 
 & RPD & RPD & RPD & $t_{r}$(s) & RPD & $t_{r}$(s) & RPD & $t_{r}$(s) \\ \hline 
Tai01$\sim$10 & 0.14 & 0.704  & 0.24  & 4.01  & 0.04  & 4.50  & 0.03  & 13.5  \\ \hline 
Tai11$\sim$20 & 0.18 & 0.843  & 0.77  & 4.09  & 0.02  & 9.00  & 0.02  & 26.3  \\ \hline 
Tai21$\sim$30 & 0.14 & 0.720  & 0.85  & 4.63  & 0.05  & 18.00  & 0.05  & 69.3  \\ \hline 
Tai31$\sim$40 & 0.06 & 0.090  & 0.12  & 6.38  & 0.00  & 11.25  & 0.00  & 2.8  \\ \hline 
Tai41$\sim$50 & 0.33 & 0.746  & 2.01  & 9.94  & 0.72  & 22.50  & 0.57  & 79.8  \\ \hline 
Tai51$\sim$60 & 1.78 & 1.855  & 3.29  & 11.82  & 0.99  & 45.00  & 1.36  & 168.1  \\ \hline 
Tai61$\sim$70 & 0.17 & 0.072  & 0.11  & 15.31  & 0.01  & 22.50  & 0.00  & 52.6  \\ \hline 
Tai71$\sim$80 & 0.33 & 0.404  & 0.66  & 18.79  & 0.16  & 45.00  & 0.18  & 211.0  \\ \hline 
Tai81$\sim$90 & 2.11 & 0.985  & 3.17  & 24.04  & 1.30  & 90.00  & 1.45  & 310.8  \\ \hline 
Average & 0.58 & 0.713 & 1.25  & 11.00  & 0.37 & 29.75 & 0.41  & 103.8  \\ \hline 
\end{tabular}
\label{Tab:RTaillardSetExisting}
\end{table}

\begin{table} [t]
\centering \caption{Results by RTGO with $N$=10, 30 and 50 on the Taillard's dataset.}
\begin{tabular}{|c|c|c|c|c|c|c|c|} \hline 
\multirow{2}{*}{Instance} & \multirow{2}{*}{$n$, $m$} & \multicolumn{2}{|c|}{RTGO($N$=10)} & \multicolumn{2}{|c|}{RTGO($N$=30)} & \multicolumn{2}{|c|}{RTGO($N$=50)} \\ \cline{3-8} 
 &  & RPD & $t_{r}$(s) & RPD & $t_{r}$(s) & RPD & $t_{r}$(s) \\ \hline 
Tai01$\sim$10 & 20,5 & 0.174  & 0.025  & 0.093  & 0.055  & 0.041  & 0.083  \\ \hline 
Tai11$\sim$20 & 20,10 & 0.285  & 0.088  & 0.102  & 0.243  & 0.033  & 0.291  \\ \hline 
Tai21$\sim$30 & 20,20 & 0.192  & 0.140  & 0.062  & 0.328  & 0.052  & 0.467  \\ \hline 
Tai31$\sim$40 & 50,5 & 0.118  & 0.068  & 0.041  & 0.175  & 0.029  & 0.276  \\ \hline 
Tai41$\sim$50 & 50,10 & 0.894  & 0.616  & 0.582  & 2.340  & 0.540  & 3.804  \\ \hline 
Tai51$\sim$60 & 50,20 & 1.296  & 1.702  & 0.903  & 6.435  & 0.747  & 11.29  \\ \hline 
Tai61$\sim$70 & 100,5 & 0.067  & 0.218  & 0.021  & 0.483  & 0.020  & 0.713  \\ \hline 
Tai71$\sim$80 & 100,10 & 0.466  & 0.995  & 0.285  & 3.597  & 0.197  & 7.087  \\ \hline 
Tai81$\sim$90 & 100,20 & 1.614  & 4.974  & 1.197  & 20.48  & 1.010  & 42.29  \\ \hline 
Average & - & 0.567 & 0.981  & 0.365 & 3.793  & 0.297 & 7.367  \\ \hline 
\end{tabular}
\label{Tab:RTaillardSetRTGO}
\end{table}

Table \ref{Tab:HellerSet} compares the results by the RTGO with $N$=30, DPSO2 \cite{Deroussi:2005p2440}, and NEH-ALA \cite{Agarwal:2006p2466} on the Heller's dataset. DPSO2 \cite{Deroussi:2005p2440} is a discrete PSO hybridized with variable neighborhood search (VNS), and was executed on a 2.4-GHz Pentium IV processor. NEH-ALA \cite{Agarwal:2006p2466} is an adaptive-learning approach in conjunction with the NEH heuristic \cite{Nawaz:1983p2441}, which is coded in Visual Basic 6.0 and was run on a 933-MHz processor. The $r_{TR} $ values of RTGO versus DPSO2 and NEH-ALA are 387.8 and 1132.3, respectively. The results indicated that RTGO performs much better than DPSO2 and NEH-ALA on Hel1 and Hel2.

Table \ref{Tab:RTaillardSetExisting}  summarizes the results by five existing algorithms, i.e., HSA \cite{Nearchou:2004p2443}, PACO \cite{Rajendran:2004p2504}, ILS \cite{Ruiz:2005p2510,Stutzle:1998p2467}, HGA\_RMA \cite{Ruiz:2006p2435}, and PSO$_{VNS}$ \cite{Tasgetiren:2007p2491} on the Taillard's dataset \cite{Taillard:1993p2472}. HSA is a population-based hybrid simulated annealing (SA). PACO is a newly-developed ant-colony optimization (ACO). ILS is an iterated local search \cite{Stutzle:1998p2467}, which is implemented by Ruiz and Maroto \cite{Ruiz:2005p2510} in Delphi 6.0 and was run on a 1.4-GHz Athlon XP processor. HGA\_RMA is a hybrid GA with the similar block order crossover (SBOX), which is coded in Delphi 7.0 and was run on a 2.8-GHz Pentium IV processor. For HGA\_RMA, the running time is defined by the expression $n\cdot (m/2)\cdot 90$. PSO$_{VNS}$ is coded in C and was run on a 2.6-GHz Pentium IV processor. Moreover, with regard to the upper bounds for calculating RPD values, HSA, PACO, and PSO$_{VNS}$ use UB93 \cite{Taillard:1993p2472}, while ILS and HGA\_RMA use UB04 \cite{Ruiz:2006p2435}. Hence, all the five algorithms in Table 5 should have larger RPD values if they use UB05.

Table \ref{Tab:RTaillardSetRTGO} gives the results by the RTGO with $N$=10, 30, and 50, respectively. The $r_{TR} $ values of RTGO with $N$=10 versus ILS, and RTGO with $N$=30 versus HGA\_RMA and PSO$_{VNS}$ are 11.2, 15.7, and 50.8, respectively. RTGO with $N$=10 performed better than HAS and PACO on RPD, and better than ILS on both RPD and running time. RTGO with $N$=30 produced better RPD and running time than both HGA\_RMA and PSO$_{VNS}$. In addition, RTGO with $N$=50 obtained better RPD than RTGO with $N$=30, although the more agents is used, the more running time is required.

\subsubsection{For QAP}
\label{sec:QAP}

In the following experiments, only the RTGO using MP(U/P/MP) is considered for comparing with some existing algorithms on the QAP benchmark datasets. 

\begin{table} [t]
\centering \caption{Results by RTGO with $N$=50 and five existing algorithms on the bur26* dataset.}
\begin{tabular}{|c|c|c|c|c|c|c|c|} \hline 
Instance & \textit{f*} & RTGO & GRASP & ANT & HAS & IFLS & GA-1 \\ \hline 
bur26a & 5426670 & 0.240 & 11.38 & 21.07 & 10 & 61.27 & 117.3 \\ \hline 
bur26b & 3817852 & 0.313 & 59.45 & 35.03 & 17 & 60.27 & 112.7 \\ \hline 
bur26c & 5426795 & 0.197 & 5.16 & 19.09 & 3.7 & 57.78 & 113.5 \\ \hline 
bur26d & 3821225 & 0.285 & 15.12 & 19.4 & 7.9 & 61.27 & 106.7 \\ \hline 
bur26e & 5386879 & 0.239 & 17.63 & 20.53 & 9.1 & 57.83 & 109.1 \\ \hline 
bur26f & 3782044 & 0.151 & 5.05 & 11.23 & 3.4 & 59.19 & 102.2 \\ \hline 
bur26g & 10117172 & 0.266 & 222.58 & 18.67 & 7.7 & 57.72 & 97.1 \\ \hline 
bur26h & 7098658 & 0.152 & 37.58 & 5.67 & 4.1 & 57.47 & 101.9 \\ \hline 
Average & - & 0.230 & 46.74 & 18.84 & 7.86 & 59.1 & 107.56 \\ \hline 
\end{tabular}
\label{Tab:bur26Set}
\end{table}

\begin{table} [t]
\centering \caption{Results by RTGO with $N$=50 and four existing algorithms on the esc* dataset.}
\begin{tabular}{|c|c|c|c|c|c|c|c|c|} \hline 
\multirow{2}{*}{Instance} & \multirow{2}{*}{\textit{f*}}  & RTGO & \multicolumn{2}{|c|}{GRASP} & IFLS & GA$_{CT}$ & \multicolumn{2}{|c|}{GA-1} \\ \cline{3-9} 
 &  & $t_{r}$ & RPD & $t_{r}$ & $t_{r}$ & $t_{r}$ & RPD & $t_{r}$ \\ \hline 
esc32a & 130 & 0.789 & 1.54 & 7.03 & 136.8 & 21.0 & 3.08 & 190.9 \\ \hline 
esc32b & 168 & 0.094 & 0.00 & 2.80 & 110.4 & 18.0 & 0.00 & 200.1 \\ \hline 
esc32c & 642 & 0.014 & 0.00 & 0.00 & 54.7 & 16.2 & 0.00 & 194.6 \\ \hline 
esc32d & 200 & 0.031 & 0.00 & 1.92 & 74.3 & 16.8 & 0.00 & 176.3 \\ \hline 
esc32e & 2 & 0.012 & 0.00 & 0.00 & 46.1 & - & 0.00 & 184.9 \\ \hline 
esc32f & 2 & 0.012 & 0.00 & 0.00 & 44.5 & - & 0.00 & 184.4 \\ \hline 
esc32g & 6 & 0.013 & 0.00 & 0.00 & 28.4 & - & 0.00 & 185.5 \\ \hline 
esc32h & 438 & 0.047 & 0.00 & 3.41 & 85.8 & 17.4 & 0.00 & 174.5 \\ \hline 
esc64a & 116 & 0.058 & - & - & 1521.7 & 183.0 & 0.00 & 1315.4 \\ \hline 
\end{tabular}
\label{Tab:escSet}
\end{table}

Different existing algorithms were applied on different instances. Thus the 50 QAPLIB instances are divided into five small datasets, i.e., bur26*, esc*, lipa*, tai*b, and misc*.

Some existing algorithms, such as GRASP \cite{Maniezzo:1999p2461}, CPTS \cite{James:2009p2442}, ANT \cite{Maniezzo:1999p2461}, ANT$_{SA}$ \cite{Demirel:2006p2520}, HAS \cite{Gambardella:1999p2521}, IFLS \cite{Ramkumar:2008p2501}, GA$_{CT}$ \cite{Drezner:2008p2459}, GA-1 \cite{Ahuja:2000p2511}, and GA$_{HRR}$ \cite{Misevicius:2003p2462}, have been applied on some QAPLIB instances. Here GRASP and CPTS are local-based search methods; ANT, ANT$_{ST}$ and HAS are ant-based systems; IFLS, GA$_{CT}$, GA-1, and GA$_{HRR}$ are genetic algorithms. Both GRASP and ANT are coded in Fortran 77 and were run on a 166-Mhz Pentium processor. CPTS, i.e., a cooperative parallel tabu search algorithm, is written in C and was run on ten 1.3-GHz Itanium processors. ANT$_{SA}$, or called AntSimulated \cite{Demirel:2006p2520}, which uses simulated annealing as its LS, is coded in C. HAS is a hybrid ant colony system. IFLS is coded in JAVA, and was run on a 2.4GHz Athlon XP processor. GA$_{CT}$ is a hybrid GA with a concentric tabu search \cite{Drezner:2008p2459}, which is coded in Fortran, and was tested on a 600-MHz Pentium III processor. GA-1 is a greedy GA. GA$_{HRR}$ is a GA hybridized with ruin and recreate procedure.

For an algorithm, the RPD column is not listed if the algorithm can achieved 100\% success rate for all the tested instances. The calculation of an $r_{TR} $ value for two algorithms is applied only on the instances tested by both the related algorithms. 

\begin{table} [t]
\centering \caption{Results by RTGO with $N$=50 and three existing algorithms on the lipa* dataset.}
\begin{tabular}{|c|c|c|c|c|c|c|c|c|} \hline 
\multirow{2}{*}{Instance} & \multirow{2}{*}{\textit{f*}} & RTGO & \multicolumn{2}{|c|}{ANT} & \multicolumn{2}{|c|}{IFLS} & \multicolumn{2}{|c|}{GA-1} \\ \cline{3-9}
 &  & $t_{r}$ & RPD & $t_{r}$ & RPD & $t_{r}$ & RPD & $t_{r}$ \\ \hline 
lipa20a & 3683 & 0.055 & 0.00 & 107.32 & 0.00 & 16.11 & 0.00 & 37.4 \\ \hline 
lipa30a & 13178 & 0.313 & 0.00 & 54.85 & 0.00 & 119.72 & 0.00 & 172.3 \\ \hline 
lipa40a & 31538 & 3.053 & 1.02 & 281.00 & 0.00 & 489.91 & 0.00 & 510.9 \\ \hline 
lipa50a & 62093 & 10.225 & - & - & 1.02 & 1556.28 & 0.95 & 743.1 \\ \hline 
lipa20b & 27076 & 0.027 & 0.00 & 0.00 & 0.00 & 16.78 & 0.00 & 37.2 \\ \hline 
lipa30b & 151426 & 0.077 & 0.00 & 0.00 & 0.00 & 121.81 & 0.00 & 168.6 \\ \hline 
lipa40b & 476581 & 0.203 & 0.00 & 0.00 & 0.00 & 485.98 & 0.00 & 513.2 \\ \hline 
lipa50b & 1210244 & 0.469 & - & - & 0.00 & 1461.81 & 0.00 & 754.7 \\ \hline 
\end{tabular}
\label{Tab:lipaSet}
\end{table}

\begin{table} [t]
\centering \caption{Results by RTGO with $N$=50 and four existing algorithms on the tai*b dataset.}
\begin{tabular}{|c|c|c|c|c|c|c|c|c|} \hline 
\multirow{2}{*}{Instance} & \multirow{2}{*}{\textit{f*}} & RTGO & CPTS & \multicolumn{2}{|c|}{ANT$_{SA}$} & \multicolumn{2}{|c|}{HAS} & GA$_{HRR}$ \\ \cline{3-9}
 &  & $t_{r}$ & $t_{r}$ & RPD & $t_{r}$ & RPD & $t_{r}$ & $t_{r}$ \\ \hline 
tai20b & 122455319 & 0.058 & 0.1 & 0.0000 & 27 & 0.0905 & 27 & 3.1 \\ \hline 
tai25b & 344355646 & 0.204 & 0.4 & 0.0000 & 50 & 0.0000 & 12 & 5.6 \\ \hline 
tai30b & 637117113 & 0.778 & 1.2 & 0.0000 & 90 & 0.0000 & 25 & 9.7 \\ \hline 
tai35b & 283315445 & 1.015 & 2.4 & 0.0376 & 147 & 0.0256 & 147 & 15 \\ \hline 
tai40b & 637250948 & 1.614 & 4.5 & 0.4872 & 240 & 0.0000 & 51 & 27 \\ \hline 
tai50b & 458821517 & 7.581 & 13.8 & 0.2475 & 480 & 0.1916 & 480 & 49 \\ \hline 
tai60b & 608215054 & 14.492 & 30.4 & 0.2258 & 855 & 0.0483 & 855 & 82 \\ \hline 
\end{tabular}
\label{Tab:taibSet}
\end{table}

\begin{table} [h]
\centering \caption{Results by RTGO with $N$=100 and three existing algorithms on the misc* dataset.}
\begin{tabular}{|c|c|c|c|c|c|c|c|c|c|} \hline 
\multirow{2}{*}{Instance} & \multirow{2}{*}{\textit{f*}} & \multicolumn{2}{|c|}{RTGO} & \multicolumn{2}{|c|}{ANT} & \multicolumn{2}{|c|}{IFLS} & \multicolumn{2}{|c|}{GA-1} \\ \cline{3-10} 
 &  & RPD & $t_{r}$ & RPD & $t_{r}$ & RPD & $t_{r}$ & RPD & $t_{r}$ \\ \hline 
chr20a   & 2192 & 0.000 & 1.544 & 0.00 & 331.20 & 4.38 & 10.95 & 0.18 & 47.3 \\ \hline 
chr20c   & 14142 & 0.000 & 0.186 & 0.00 & 29.49 & 0.00 & 13.55 & 4.72 & 48.9 \\ \hline 
chr22a   & 6156 & 0.000 & 1.077 & 0.00 & 314.68 & 0.88 & 19.11 & 0.62 & 72.9 \\ \hline 
chr22b   & 6194 & 0.000 & 1.770 & 0.97 & 161.75 & 1.68 & 17.00 & 1.19 & 76.1 \\ \hline 
chr25a   & 3796 & 0.000 & 1.264 & 0.00 & 236.29 & 11.17 & 33.59 & 10.54 & 96.8 \\ \hline 
had20   & 6922 & 0.000 & 0.039 & 0.00 & 158.71 & 0.00 & 10.58 & - & - \\ \hline 
kra30a   & 88900 & 0.000 & 0.807 & 0.00 & 199.06 & 1.34 & 105.55 & 1.34 & 150.7 \\ \hline 
kra30b   & 91420 & 0.000 & 1.731 & 0.00 & 140.02 & 0.13 & 101.83 & 0.18 & 165.3 \\ \hline 
mc33 & 339416 & 0.000 & 1.019 & 0.00 & 379.65 & - & - & - & - \\ \hline 
nug20 & 2570 & 0.000 & 0.109 & 0.00 & 119.28 & 0.00 & 16.06 & 0.00 & 48.9 \\ \hline 
nug30 & 6124 & 0.000 & 1.610 & 0.00 & 180.75 & 2.12 & 116.66 & 0.07 & 177.1 \\ \hline 
rou20 & 725522 & 0.002 & 2.171 & 0.00 & 244.54 & 0.02 & 11.73 & 0.08 & 37.6 \\ \hline 
scr20 & 110030 & 0.000 & 0.190 & 0.00 & 46.09 & 0.00 & 12.69 & 0.03 & 39.8 \\ \hline 
sko42 & 15812 & 0.000 & 8.458 & - & - & 0.30 & 613.92 & 0.23 & 503.1 \\ \hline 
ste36a   & 9526 & 0.000 & 7.249 & 0.76 & 295.23 & 0.00 & 204.36 & 1.47 & 354.8 \\ \hline 
ste36b   & 15852 & 0.000 & 1.319 & 0.25 & 212.81 & 3.43 & 222.45 & - & - \\ \hline 
tho30 & 149936 & 0.000 & 1.400 & 0.00 & 287.50 & 0.29 & 118.94 & 0.31 & 197.8 \\ \hline 
tho40 & 240516 & 0.008 & 26.372 & 0.66 & 312.46 & 0.53 & 501.77 & 0.33 & 479.1 \\ \hline 
\end{tabular}
\label{Tab:miscSet}
\end{table}

Table \ref{Tab:bur26Set} compares the results by RTGO with $N$=50, GRASP \cite{Maniezzo:1999p2461}, ANT \cite{Maniezzo:1999p2461}, HAS \cite{Gambardella:1999p2521}, IFLS \cite{Ramkumar:2008p2501}, and GA-1 \cite{Ahuja:2000p2511} on the bur26* dataset. Only the running times are listed, since all these algorithms achieved 100\% success rate. The $r_{TR} $ values of RTGO versus the five algorithms are 24.1, 9.7, 34.2, 440.5, and 467.7, respectively.

Table \ref{Tab:escSet} shows the results by RTGO with $N$=50, GRASP \cite{Maniezzo:1999p2461}, IFLS \cite{Ramkumar:2008p2501}, GA$_{CT}$ \cite{Drezner:2008p2459}, and GA-1 \cite{Ahuja:2000p2511} on the esc* dataset. Here RTGO, IFLS, and GA$_{CT}$ obtained a 100\% success rate, whereas GRASP and GA-1 could not fully solve esc32a. The $r_{TR} $ values of RTGO versus the four algorithms are 1.8, 3368.8, 113.0, and 2623.0, respectively. For $r_{t}$, RTGO was comparable to GRASP, but much faster than IFLS, GA$_{CT}$, and GA-l, on the esc* dataset.

Table \ref{Tab:lipaSet} compares the results by the RTGO with $N$=50, ANT \cite{Maniezzo:1999p2461}, IFLS \cite{Ramkumar:2008p2501}, and GA-1 \cite{Ahuja:2000p2511} on the lipa* dataset. Here only RTGO obtained a 100\% success rate. The $r_{TR} $ values of RTGO versus the four algorithms are 14.1, 507.4, and 203.7, respectively. 

Table \ref{Tab:taibSet} shows the results by RTGO with $N$=50, CPTS \cite{James:2009p2442}, ANT$_{SA}$ \cite{Demirel:2006p2520}, HAS \cite{Gambardella:1999p2521}, and GA$_{HRR}$ \cite{Misevicius:2003p2462} on the tai*b dataset. Here RTGO, CPTS, and GA$_{HRR}$ obtained a 100\% success rate. The $r_{TR}$ values of RTGO versus the four algorithms are 1.9, 73.4, 62.0, and 7.4, respectively. RTGO was comparable with CPTS, but much faster than ANT$_{SA}$ and HAS. Moreover, it should be noticed that CPTS is a parallel algorithm with ten processors. Naturally, the performance of RTGO can also be improved significantly by assigning agents to different processors due to the inherent parallelism in a cooperative group. 

Table \ref{Tab:miscSet} gives the results by RTGO with $N$=100, ANT \cite{Maniezzo:1999p2461}, IFLS \cite{Ramkumar:2008p2501}, and GA-1 \cite{Ahuja:2000p2511} on the misc* dataset. The $r_{TR} $ values of RTGO versus the three algorithms are 8.7, 63.8, and 44.6, respectively. Furthermore, RTGO achieved a 100\% success rate for all 18 instances except for rou20 and tho40. In terms of RPD, RTGO also performed much better. 

\subsection{Discussion}

Due to its simplicity, RTGO might be interpreted straightforwardly from the viewpoint of an agent. Over sessions, each agent can be seen as walking in the rugged problem landscape, and the current location is its base idea (its best-so-far solution). Each stable LS strategy can reach to a (near) local optimum but cannot escape from the current one, whereas the round-table group support mechanism can only provides diverse social ideas. Thus only the XS strategy provides the agent a capability of escaping from local minima. With an SBX operator, the base idea is adaptively modified by the social and repair policies using the guidance clues in the social idea. In contrast, iterated local search (ILS) \cite{Stutzle:1998p2467, Ruiz:2007p2496} can only use blind perturbations to have a randomized walk in the space of local optima. RTGO might also benefit from the cooperative portfolio effect \cite{Gomes:2001p2488} on the agents. Compared to genetic local search \cite{Xu2011} and memetic algorithms (MA) \cite{Merz:2000p2516}, each agent in RTGO has its private memory, which allow a natural way to perserve diverse ideas for possible improvements.

As shown in Section \ref{sec:IdeaComb}, good performance can be achieved by using appropriate SBX operators, for different sequencing problems. The modular design of SBX might help us to roughly identify some structure features of different problems. FSP and QAP respectively prefer more precedence and position information. The common-avoiding style does help on the 1P base policy, and the parallel macro policy has more usage on FSP that on QAP.

From the results shown in Section \ref{sec:ExistingAlgo}, RTGO can achieve better performance than many existing methods, including some metaheuristic methods that using the same LS strategies, e.g., HGA\_RMA \cite{Ruiz:2006p2435} and ILS \cite{Stutzle:1998p2467} for FSP, and HAS \cite{Gambardella:1999p2521} and IFLS \cite{Ramkumar:2008p2501} for QAP. 

RTGO is a preliminary step in studying and utilizing the group creativity from the viewpoint of a metaheuristic framework. We have shown that group creativity can emerge from a group of agents with very limited memory and thinking capability. The research in this direction might help for unravelling the real-world complexity, given that large-scale experiments in human groups might be too expensive and might have too much uncertainty. 

\section{Conclusion}
\label{sec:Conclusion}

The round-table group optimizer (RTGO) is a very simple realization of a cooperative idea-generating group. The group-level support is a round-table mechanism for providing one social idea for each agent, and each agent only stores the best-so-far idea as the base idea in its individual memory. Given the base and social ideas, the idea combination and local improvement processes of each agent are respectively realized by using a XS strategy and a LS strategy, in the form of socially biased learning. The agents are able to search in the problem landscape in parallel based on their individual memory, as well as utilize the stimulation from social ideas for achieving a collective performance, over iterative sessions.

The implementation of RTGO was performed on two sequencing problems, i.e., FSP and QAP. Two stable, domain-specific LS strategies were adopted from existing algorithms. Then a general XS class, called socially biased combination (SBX), was realized in a modular form: A basic SBX operator contains three policies, i.e., base, social and repair policies, and a macro SBX operator can be realized by applying a macro policy on any SBX operator(s).

We then evaluated the performance of the RTGO metaheuristic framework on some commonly-used FSP and QAP benchmark datasets. The effects of idea combination processes were evaluated on different SBX realizations. RTGO outperformed many existing methods, including some methods that using the same LS strategies, in terms of the solution quality and the running time. The results might also indicate that appropriate SBX operators have the capability of leaping adaptively in the problem landscape, while the diversity of promising ideas can be well-preserved by the simple RTGO framework.

There are several aspects of the proposed method that warrant further study. One issue concerns the development of more effective search strategies. For example, adaptive memory strategies \cite{Yin2010} rather than a stable LS strategy might be used to overcome local minima. The group metaphor is very open for efficient computation using any sophisticated operators. 

Moreover, RTGO variants might be designed for promoting group creativity. For example, each agent might possess a pool of ideas in its  memory \cite{Kohn2011}, so that the agent has a strong individual learning capability and can adaptively using the base and social ideas in different sessions. Each agent might also hold multiple search strategies that can be automatic configured by a hands-off learning procedure for handling complex problem domains.

\bibliographystyle{model1b-num-names}
\bibliography{bib/permutation,bib/own,bib/group}
\end{document}